\documentclass[Afour,sageh,times]{sagej}

\usepackage{moreverb,url}
\usepackage[colorlinks,bookmarksopen,bookmarksnumbered,citecolor=red,urlcolor=red]{hyperref}
\usepackage{amssymb}
\usepackage{amsmath}
\usepackage{graphicx}
\usepackage{caption}
\usepackage{subcaption}
\usepackage{algorithm}
\usepackage{algorithmicx}
\usepackage{algpseudocode}

\usepackage{xcolor}

\newcommand\BibTeX{{\rmfamily B\kern-.05em \textsc{i\kern-.025em b}\kern-.08em
T\kern-.1667em\lower.7ex\hbox{E}\kern-.125emX}}

\def\volumeyear{2016}

\begin{document}

\runninghead{Broad et al.}

\title{Data-driven Koopman Operators for Model-based Shared Control of Human-Machine Systems}

\author{Alexander Broad\affilnum{1,2,4}, Ian Abraham\affilnum{3}, Todd Murphey\affilnum{3} and Brenna Argall\affilnum{2,3,4}}

\affiliation{\affilnum{1}Boston Dynamics, Waltham, MA 02451\\
\affilnum{2}Department of Electrical Engineering and Computer Science, Northwestern University, Evanston, IL 60208\\
\affilnum{3}Department of Mechanical Engineering, Northwestern University, Evanston, IL 60208\\
\affilnum{4}Shirley Ryan AbilityLab, Chicago, IL 60611}

\corrauth{Alexander Broad,
Northwestern University, 
Evanston, IL 
60208, U.S.A.}

\email{alex.broad@u.northwestern.edu}

\begin{abstract}
We present a data-driven shared control algorithm that can be used to improve a human operator's control of complex dynamic machines and achieve tasks that would otherwise be challenging, or impossible, for the user on their own.  Our method assumes no \textit{a priori} knowledge of the system dynamics. Instead, both the dynamics and information about the user's interaction are learned from observation through the use of a Koopman operator.  Using the learned model, we define an optimization problem to compute the autonomous partner's control policy. Finally, we dynamically allocate control authority to each partner based on a comparison of the user input and the autonomously generated control.  We refer to this idea as model-based shared control (MbSC).  We evaluate the efficacy of our approach with two human subjects studies consisting of 32 total participants (16 subjects in each study).  The first study imposes a linear constraint on the modeling and autonomous policy generation algorithms.  The second study explores the more general, nonlinear variant.  Overall, we find that model-based shared control significantly improves task and control metrics when compared to a natural learning, or user only, control paradigm.  Our experiments suggest that models learned via the Koopman operator generalize across users, indicating that it is not necessary to collect data from each individual user before providing assistance with MbSC.  We also demonstrate the data-efficiency of MbSC and consequently, it's usefulness in online learning paradigms.  Finally, we find that the nonlinear variant has a greater impact on a user's ability to successfully achieve a defined task than the linear variant.
\end{abstract}


\maketitle

\section{Introduction}
\label{sec-introduction}

\begin{figure}[t]
	\centering
	\includegraphics[width=0.95\hsize]{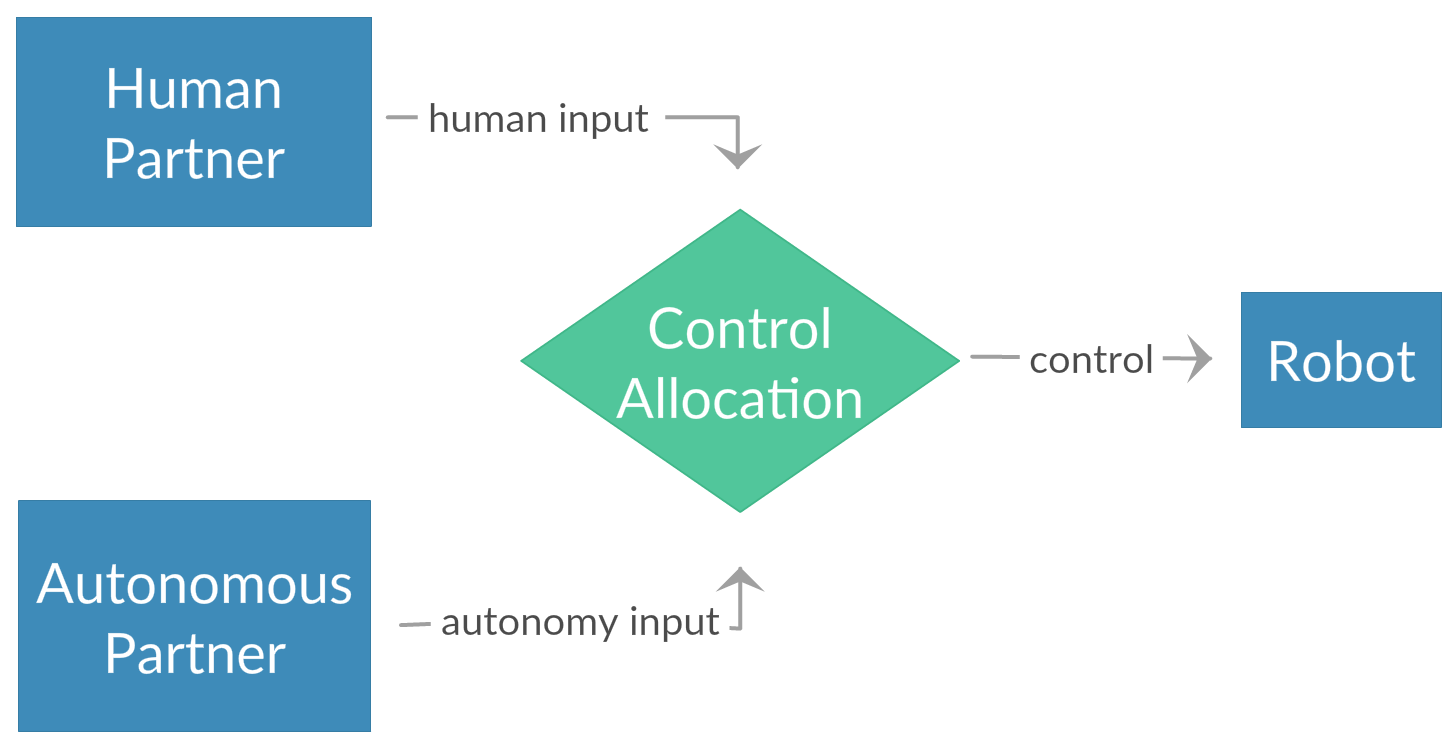}
	\caption{Pictorial representation of a shared control paradigm.  Both the human and autonomy are capable of controlling the mechanical system, and a dynamic control allocation algorithm selects which agent is in control at any given moment.}
	\label{fig-shared-control}
\end{figure}

Robot autonomy offers great promise as a tool by which we can enhance, or restore, the natural abilities of a human partner.  For example, in the fields of assistive and rehabilitative medicine, devices such as exoskeletons and powered wheelchairs can be used to assist a human who has severely diminished motor capabilities.  However, many assistive devices can be difficult to control.  This can be due to the inherent complexity of the system, the required fidelity in the control signal, or the physical limitations of the human partner.  We can, therefore, further improve the efficacy of these devices by offloading challenging aspects of the control problem to an autonomous partner.  In doing so, the human operator is freed to focus their mental and physical capacities on important high-level tasks like path planning and interaction with the environment.  This idea forms the basis of \textit{shared control} (see Figure~\ref{fig-shared-control}), a paradigm that aims to produce joint human-machine systems that are more capable than either the human or machine on their own.  

A primary challenge that researchers and engineers face when developing shared control paradigms for generic human-machine systems is a lack of \textit{a priori} knowledge of the human and robot partners.  This issue is compounded by the fact that, in the real world, many users may operate the same mechanical device.  It is therefore necessary to consider solutions that generalize to a variety of potential human and machine partners.  In this work, we propose a data-driven methodology that learns all relevant information about how a given human and machine pair interact directly from observation.  We then integrate the learned model of the joint system into a single shared control paradigm.  We refer to this idea as \textit{model-based shared control}.

In this work, we learn a model of the joint human-machine system through an approximation to the Koopman operator (\cite{koopman1931hamiltonian}), though any machine learning approach could be used.  However, the Koopman operator is chosen specifically for this work as it has previously proven useful in human-in-the-loop systems~(\cite{broad2017learning}) and can be computed efficiently~(\cite{williams2015kernel}). This model is trained on observation data collected during demonstration of the human and machine interacting and therefore describes both the human's input to the system, and the robot's response to the human input and system state.  We can then integrate the portion of the learned model that specifically describes the system and control dynamics of the mechanical device into an optimal control algorithm to produce autonomous policies.  Finally, the input provided by the human and autonomous partners are integrated via a geometric signal filter to provide real-time, dynamic shared control of unknown systems.  

We validate our thesis that modeling the joint human-machine system is sufficient for the purpose of automating assistance with two human subjects studies consisting of 32 total participants.  The first study imposes a linear constraint on the modeling and control algorithms, while the second study relaxes these constraints to evaluate the more general, nonlinear case.  The linear variant of our proposed algorithm is used to validate the efficacy of our shared control paradigm and was first presented in~\cite{broad2017learning}.  The nonlinear variant extends these results to a wider class of human-machine systems.  The results of the two studies demonstrate that the nonlinear variant has a greater impact on overall task performance than the linear methods.  We also find that our modeling technique is generalizable across users with results that suggest that individualizing the model offline, based on a user's own data, does not affect the ability to learn a useful representation of the dynamical system.  Finally, we evaluate the efficacy of our shared control paradigm in an online learning scenario, demonstrating the sample efficiency of the model-based shared control paradigm.

We provide background and related work in Section~\ref{sec-background-and-related-work}.  We then define model-based shared control in Section~\ref{sec-model-based-shared-control}.  In Section~\ref{sec-experimental-validation} we describe the human subjects study we perform and detail the results in Section~\ref{sec-results}.  We describe important takeaways in Section~\ref{sec-discussion} and conclude in Section~\ref{sec-conclusion}.

\section{Background and Related Work}
\label{sec-background-and-related-work}

This section presents background and related work in the shared control literature for human-machine systems.  We also identify alternative methods of autonomous policy generation for shared control, and provide a detailed background on the Koopman operator (\cite{koopman1931hamiltonian}) with a particular focus on its use in learning system dynamics.

\subsection{Shared Control}
\label{sec-background-shared-control}

In this work, we explore the question of how automation can be used to adjust to, and account for, the specific capabilities of a human partner.  In particular, we aim to develop a methodology that allows us to \textit{dynamically adjust} the amount of control authority given to the robot and human partners~(\cite{hoeniger1998dynamically, hoffman2004collaboration}).  If done intelligently, and with appropriate knowledge of the individual capabilities of each team member, we can improve the overall efficiency, stability and safety of the joint system~(\cite{lasota2017survey}).  Approaches to shared control range from pre-defined, discretely adjustable methods~(\cite{kortenkamp2000adjustable}) to probabilistic models~(\cite{javdani2015shared}) to policy blending~(\cite{dragan2013policy}).  In addition to blending in the original control signal space, shared control has been researched through haptic control~(\cite{nudehi2005shared}) and compliant control~(\cite{kim1992force}).

In this work, we allocate control using a filter~(\cite{tzorakoleftherakis2015controllers}) described more thoroughly in Section~\ref{sub-sec-control-allocation}.  Our control allocation strategy is similar in practice to \textit{virtual fixtures} and \textit{virtual guides}, techniques that are common in the haptics literature~(\cite{forsyth2005predictive, griffiths2005sharing}).  In particular, virtual fixtures and guides are techniques by which autonomously generated forces are \textit{added to the control of a system} to limit movement into undesriable areas and/or influence motion towards an optimal strategy~(\cite{abbink2012haptic}).  These ideas have been explored most commonly in association with robotic telemanipulation~(\cite{abbott2007haptic}), including applications like robotic surgery~(\cite{marayong2004speed}) and robot-assisted therapy~(\cite{noohi2016model}).  A key difference between these approaches and our own is that our control allocation method does not incorporate additional information from the autonomous partner into the control loop.  Instead, the autonomous partner simply rejects input from the operator that does not meet the proposed criteria. Our approach therefore requires no \textit{a priori} information about (or ability to sense) the environment, and no information about the system dynamics. In contrast, virtual fixtures/guides require information about (or the ability to detect) hard constraints in the environment, and knowledge of the system dynamics. This information is then used to compute forces---the virtual fixtures---that  counteract user-generated forces that are defined by as dangerous.  The approach in this paper does not have similar \textit{a priori} information requirements, suggesting our approach can more easily be incorporated into novel human-machine systems. An important benefit of the methods proposed in the virtual fixtures/guide literature is that the techniques often provide an explicit guarantee of safety for the joint human-machine system. Our approach can be extended to provide the same guarantees by incorporating information about (or the ability to sense) the environment and using control barrier functions to implement safety requirements~(\cite{broad2018operation}).

The effects of shared control (SC) have been explored in numerous fields in which the addition of a robot partner could benefit a human operator.  For example, in assistive and rehabilitation robotics, researchers have explored the effects of shared control on teleoperation of smart wheelchairs~(\cite{erdogan2017effect, trieu2008shared}) and robotic manipulators~(\cite{kim2006continuous}).  Similarly, researchers have explored shared control as it applies to the teleoperation of larger mobile robots and human-machine systems, such as cars~(\cite{dewinter11smc}) and aircraft~(\cite{matni08acc}).  When dealing with systems of this size, safety is often a primary concern.  

The above works are conceptually similar to our own as they use automation to facilitate control of a robot by a human partner.  However, in this work, we do not augment the user's control based on an explicit model of the user.  Instead, we use observations of the user demonstrations to build a \textit{model of the joint human-robot system}.  The effect of the human partner on the shared control system is implicitly encoded in the model learned from their interactions.

\subsection{Model-Based Reinforcement Learning}
\label{sec-background-model-based}

Model-based shared control (MbSC) is a paradigm that generalizes shared control to generic human-machine partners~(\cite{broad2017learning}).  That is, MbSC assumes no \textit{a priori} knowledge of either partner and instead uses data-driven techniques to learn models of the human and/or robot partner(s) from observation.  In addition to providing a quantitative understanding of each partner, these models can be used to generate autonomous control policies by integrating the learned system and control dynamics into an optimal control framework.  

Model-based shared control is therefore highly related to model-based reinforcement learning (MbRL), a paradigm that explicitly learns a model of the system dynamics in addition to learning an effective control policy.  MbSC extends MbRL to systems that integrate control from various sources.  Early work in model-based reinforcement learning includes~(\cite{barto1995learning}) and~(\cite{kaelbling1996reinforcement}).  More recently, researchers have considered integrating learned system models with optimal control algorithms to produce control trajectories in a more data-efficient manner~(\cite{mitrovic2010adaptive}).  These algorithms compute control through an online optimization process, instead of through further data collection~(\cite{barto1995learning}).  There are of course, many viable model learning techniques that can be used to describe the system and control dynamics.  For example, Neural Networks~(\cite{williams2017information}), Gaussian Processes~(\cite{nguyen2009local}), and Gaussian Mixture Models~(\cite{khansari2011learning}) have all shown great promise in this area.  Often the best choice of modeling algorithm is related specifically to the application domain.  For example, Gaussian Processes perform well in low-data regimes, but scale poorly with the size of the dataset where Neural Networks fit naturally.  In this work we explore a modeling technique that easily integrates with derivative-based optimal control algorithms.  A survey of learning for control can be found in~(\cite{schaal2010learning}).

From a motivational standpoint, related work also includes methods that model not only the dynamics of a robotic system, but combined human-machine systems from data. For example, researchers have explored learning control policies from user demonstrations, thereby incorporating both system dynamics and the user's desires~(\cite{argall2009survey, celemin2019fast}).  Building on these ideas, researchers have proposed learning shared control policies directly from demonstration data using deep reinforcement learning~(\cite{reddy2018shared}).  To improve the human partner's intuition for the interaction paradigm, researchers have also proposed learning latent spaces to allow users to control complex robots with low dimensional input devices~(\cite{losey2019controlling}).  Relatedly, people have also proposed techniques for modeling both the dynamics of a system, and a policy for deciding when a human or autonomous partner should be in control.  One such method is to learn local approximations to the system's dynamics and only provide autonomous assistance when the system is nearby a state it has previously observed~(\cite{peternel2016shared}).  Our approach utilizes a linear representation of the nonlinear human-robot dynamics which avoids the use of local models in exchange for a higher capacity linear model which globally represents the complex system.  This is also distinct from the virtual fixtures/guide literature where system models are known \textit{a priori}, and frequently nonlinear.

From a methodological standpoint, the most closely related research is recent work that computes control trajectories by integrating learned dynamics models with model predictive control (MPC) algorithms~(\cite{williams2017information, drews2017aggressive}).  These algorithms are defined by an iterative, receding horizon optimization process instead of using an infinite-horizon.  Similar to our own work, these researchers first collect observations from live demonstrations of the mechanical device to learn a model of the system dynamics.  They then integrate the model with an MPC algorithm to develop control policies.  Beyond methodological differences (e.g., choice of machine learning and optimal control algorithms), the key theoretical distinction between these works and our own is our focus on shared control of joint human-machine systems, instead of developing fully autonomous systems.  In particular, we learn a model of the joint system that is integrated into a shared control system to improve a human operator's control of a dynamic system.  We therefore consider the influence of the human operator both during the data-collection process and at run-time in the control of the dynamic system.

In this work, we learn a model of the system and control dynamics through an approximation to the Koopman operator~(\cite{koopman1931hamiltonian}).  As the Koopman operator is a relatively new concept in robot learning for control, we now provide additional information on its description in the following section.

\subsection{The Koopman Operator}
\label{sec-background-koopman}

\begin{figure*}[!th]
	\centering
	\includegraphics[width=0.8\hsize]{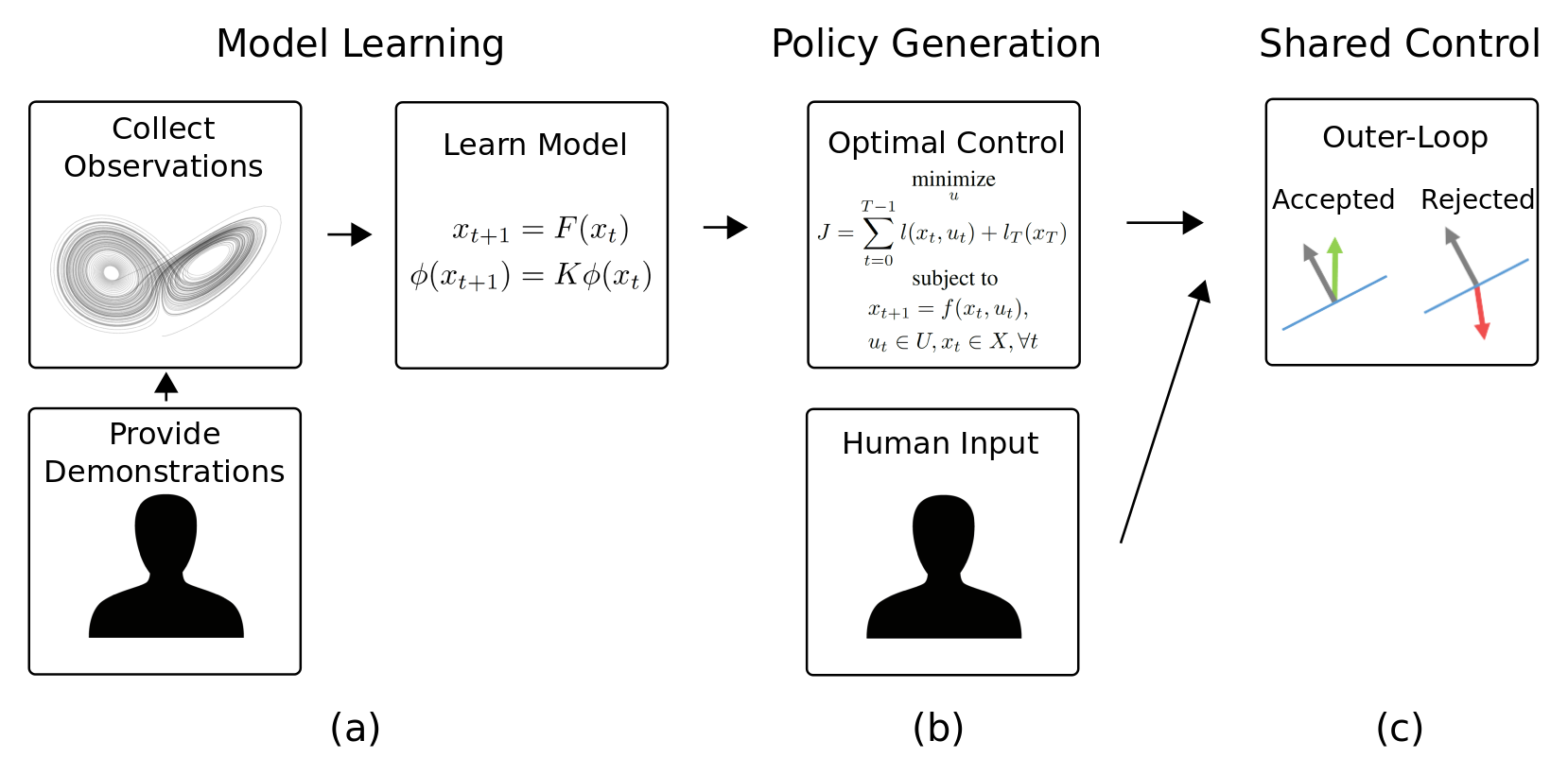}
	\caption{Pictorial depiction of the our model-based shared control paradigm.  (a) Collect observations from user interaction and learn a model of the joint human-machine system through an approximation to the Koopman operator.  This can be computed offline or online.  (b) Compute control policy of autonomous agent by solving optimal control problem using the learned model.  (c) Allocate control to integrate autonomy (gray) and user input (green/red).}
	\label{fig-pipeline}
\end{figure*}

The Koopman operator is an infinite-dimensional linear operator that can capture all information about the evolution of nonlinear dynamical systems.  This is possible because the operator describes a linear mapping between sequential \textit{functions of states} instead of the state itself.  In particular, the Koopman operator acts on an infinite dimensional Hilbert space representation of the state.  To define the Koopman operator, let us consider a discrete time dynamic system $(\mathcal{X}, t, F)$: 
\begin{equation}
x_{t+1} = F(x_t)
\label{eq-gen-dynamics}
\end{equation}

\noindent where $\mathcal{X} \subseteq \mathbb{R}^N$ is the state space, $t \in \mathbb{R}$ is time and $F : \mathcal{X} \rightarrow \mathcal{X}$ is the state evolution operator.  We also define $\phi$, a nominally infinite dimensional observation function 
\begin{equation}
y_t = \phi(x_t)
\label{fn-obs}
\end{equation}

\noindent where $\phi : \mathcal{X} \rightarrow \mathbb{C}$ defines the transformation from the original state space into the Hilbert space representation that the Koopman operator acts on.  The Koopman operator $\mathcal{K}$ is defined as the composition of $\phi$ with $F$, such that 

\begin{equation}
\mathcal{K} \phi = \phi \circ F.
\label{fn-koopman-eq}
\end{equation}

\noindent By acting on the Hilbert state representation, the \textit{linear} Koopman operator is able to capture the complex, nonlinear dynamics described by the state evolution operator. 

While the Koopman operator is nominally infinite dimensional, recent work has demonstrated the ability to approximate a finite dimensional representation using data-driven techniques~(\cite{rowley2009spectral, budivsic2012applied}).  In the limit of collected observation data, the approximation to the Koopman becomes exact~(\cite{williams2015data}).  These data-driven methods have renewed an interest in using the Koopman operator in applied engineering fields.  In contemporary work, the Koopman operator has been successfully used to learn the dynamics of numerous challenging systems.  This includes demonstrations that show the Koopman operator can differentiate between cyclic and non-cyclic stochastic signals in stock market data~(\cite{hua2016using}) and that it can detect specific signals in neural data that signify non-rapid eye movement (NREM) sleep~(\cite{brunton2016extracting}).  More recently these systems have included physical robotics systems~(\cite{abraham2019active, bruder2019modeling}).

\section{Model-based Shared Control}
\label{sec-model-based-shared-control}

Our primary goal is to develop a shared control methodology that improves the skill of human-machine systems without relying on \textit{a priori} knowledge of the relationship between the human and the machine.  To define our model-based shared control algorithm we now describe the (1) model learning process, (2) method for computing the policy of the autonomous agent (\textit{autonomy input} in Figure~\ref{fig-shared-control}) and (3) control allocation method (the \textit{green box} in Figure~\ref{fig-shared-control}).  A pictorial depiction of our model-based shared control paradigm can be found in Figure~\ref{fig-pipeline}.  Our learning-based approach develops a model of the joint human-machine system solely from observation, and this model can be used by the policy generation method to develop autonomous control trajectories.  The control allocation method then describes how we integrate the input provided by the human partner and the autonomous agent into a single command that is sent to the dynamic system.  

\subsection{Model Learning via the Koopman Operator}

When designing assistive shared control systems, it is important to consider both the human and autonomous partners.  To ensure that our paradigm is valid for generic human-machine systems, we learn both the \textit{system dynamics} and information about the \textit{user interaction} directly from data.  In this work, we develop a model of the joint human-machine through an approximation to the Koopman operator, which can be computed offline or online (discussed further in Section~\ref{sec-study-two-results-online}).  The model learning process is depicted in Figure~\ref{fig-pipeline}\textcolor{red}{(a)}.  As mentioned previously, there are of course a variety of other machine learning algorithms and representions one could choose to learn the system dynamics.  In this work, we use the Koopman operator, which is particularly well suited to model-based shared control of human-machine systems for two main reasons.  First, it is possible to approximate the Koopman operator in low-data regimes (see Section~\ref{sec-study-two-results-online}) which allows us to quickly expand the set of human-machine systems we can control under the general MbSC paradigm.  Second, there are a variety of highly efficient learning algorithms~(\cite{williams2015data, klus2015numerical, rowley2009spectral}) that make the Koopman operator well suited to an online learning paradigm, an important feature in shared control where it is unlikely that we have \textit{a priori} knowledge of the joint human-machine system.

We use Extended Dynamic Mode Decomposition (EDMD) to approximate the Koopman operator~(\cite{williams2015data}).  EDMD belongs to a class of data-driven techniques known as Dynamic Mode Decomposition (DMD)~(\cite{rowley2009spectral, schmid2010dynamic, tu2013dynamic}).  These algorithms use snapshots of observation data to approximate the Koopman modes that describe the dynamics of the observed quantities.  We now provide a mathematical treatment of the EDMD algorithm.  We start by redefining the observation function $\phi$ from Equation~\ref{fn-obs} as a vector valued set of basis functions chosen to compute a finite approximation to the Hilbert space representation.  We can then define the following approximation to the Koopman operator

\begin{equation}
\phi(x_{t+1}) = \mathcal{K}^T\phi(x_t) + r(x_t)
\end{equation}

\noindent where $r(x_t)$ is a residual term that represents the error in the model.  The Koopman operator is therefore the solution to the optimization problem that minimizes this residual error term 
\begin{align}
\begin{split}
J & = \frac{1}{2} \sum_{t=1}^{T} |r(x_t)|^2\\
& = \frac{1}{2} \sum_{t=1}^{T} |\phi(x_{t+1}) - \phi(x_t)K) |^2
\end{split}
\label{eq-ls}
\end{align}

\noindent where $T$ is the time horizon of the optimization procedure, and $|\cdot|$ is the absolute value.  The solution to the least squares problem presented in Equation~\eqref{eq-ls} is
\begin{equation*}
K = G^\dagger A
\end{equation*}

\noindent where $\dagger$ denotes the Moore-Penrose pseudo inverse and
\begin{align*}
G & = \frac{1}{T} \sum_{t=1}^{T} \phi(x_t)^T\phi(x_t) \\
A & = \frac{1}{T} \sum_{t=1}^{T} \phi(x_t)^T\phi(x_{t+1})
\end{align*}

\subsubsection{Basis} In this work, we require that the finite basis $\phi$ \textit{includes both the state and control variables}~(\cite{proctor2016generalizing}).  This ensures that the Koopman operator models both the natural dynamics of the mechanical system and the control dynamics as provided by the user demonstration.   In this work we empirically select a fixed set of basis functions to ensure that all models (across the different users in our validation study) are learned using the same basis.  Here we choose $\phi$ such that
\begin{align}
\begin{split}
\phi = &[1, x_1, x_2, x_3, x_4, x_5, x_6, u_1, u_2, u_1*x_1, u_1*x_2, \\
& u_1*x_3, u_1*x_4, u_1*x_5, u_1*x_6, u_2*x_1, u_2*x_2, \\
& u_2*x_3, u_2*x_4, u_2*x_5, u_2*x_6, u_1*cos(x_3), \\
& u_1*sin(x_3), u_2*cos(x_3), u_2*sin(x_3)].
\end{split}
\label{eq-basis}
\end{align}
These 25 basis functions were chosen to combine information about the geometry of the task (e.g., the trigonometric functions capture specific nonlinearities present in the system dynamics, see Section~\ref{sec-experimental-environment}) with information related to how the user responds to system state. For this reason, we include terms that mix state information with control information.  To evaluate the accuracy of the learned approximation to the Koopman operator we compute the H-step prediction accuracy (see Figure~\ref{fig-h-step-accuracy}).

There are, of course, a variety of methods that one could use to select an appropriate basis for a given dynamical system.  This step is particularly important as selecting a poor basis will quickly degrade the validity of the learned model~(\cite{berrueta2018dynamical}).  One such method is to integrate known information about the system dynamics into the chosen basis functions, such as the relationship between the heading of the lander and the motion generated by the main thruster.  This approach works well when the system dynamics are easy to understand, however it can prove challenging when the dynamics are more complex.  For this reason, one could also choose the set of basis functions through purely data-driven techniques.  Sparsity Promoting DMD~(\cite{jovanovic2014sparsity}) is one such algorithm.  SP-DMD takes a large initial set of randomly generated basis functions and imposes an $\ell_1$ penalty during the learning process to algorithmically decides which basis functions are the most relevant to the observable dynamics~(\cite{tibshirani1996regression}). An example of this purely data-driven approach being applied to human-machine systems can be found in~\cite{broad2019highly}.

\subsection{Autonomous Policy Generation}

To generate an autonomous control policy, we can integrate the portion of the learned model that relates to the system and control dynamics into a model predictive control (MPC) algorithm.  In particular, we use Koopman operator model-based control~(\cite{broad2017learning, abraham2017model}), which we detail now in full.  To compute the optimal control sequence, $u$, we must solve the following Model Predictive Control (MPC) problem
\begin{equation}
\begin{aligned}
& \underset{u}{\text{minimize}}
& & J = \sum_{t=0}^{T-1} l(x_t,u_t) + l_T(x_T) \\
& \text{subject to}
& & x_{t+1} = f(x_t, u_t), \\
& & & u_t \in U, x_t \in X, \forall t
\end{aligned}
\label{eqn-mpc}
\end{equation}

\noindent where $f(x_t,u_t)$ is the system dynamics, $l$ and $l_T$ are the running and terminal cost, and $U$ and $X$ are the set of valid control and state values, respectively.

In this work, we define
\begin{align*}
l(x_t, u_t) &= \frac{1}{2} (x_t-x_d) Q_t (x_t-x_d) + \frac{1}{2} u_t R_t u_t \\
\textnormal{where }Q_t &= Diag[6.0, 10.0, 20.0, 2.0, 2.0, 3.0]
\end{align*}
$x_t$ is the current state and $x_d$ is the desired goal state.  Additionally,
\begin{align*}
l_T(x_T) &= \frac{1}{2} (x_t-x_d) Q_T (x_t-x_d) \\
\textnormal{where }Q_T &= Diag[3.0, 3.0, 5.0, 1.0, 1.0, 1.0]
\end{align*}
These values were chosen empirically based on results observed from the system operating fully autonomously.

To integrate our learned system model, we re-write the system dynamics as such:
\begin{equation}
\phi(x_{t+1}) = f_\mathcal{K}(x_t, u_t)
\label{eqn-koopman-dyn}
\end{equation}

\noindent where $f_\mathcal{K} = \mathcal{K}^T\phi(x_t, u_t)$ is the learned system dynamics parameterized by a Koopman operator $\mathcal{K}$.  This equation demonstrates the fact that the Koopman operator does not map directly from state to state, but rather operates on functions of state.  We can then evaluate the evolved state by recovering the portion of the basis that represents the system's state
\begin{equation}
x_{t+1} = \phi(x_{t+t})_{1:N}
\label{eqn-koopman-recover}
\end{equation}
\noindent where values $1:N$ are the state variables, as per our definition in Equation~\eqref{eq-basis}, and $N$ is the dimension of the state space.  The policy generation process is depicted in Figure~\ref{fig-pipeline}\textcolor{red}{(b)}.

\subsubsection{Nonlinear Model Predictive Control Algorithm}

We solve Equation~\eqref{eqn-mpc} with Sequential Action Control~(\cite{ansari2016sequential}) (SAC).   SAC is a real-time, model-based non-linear optimal control algorithm that is designed to iteratively find a single value, a time to act, and a duration that maximally improves performance.  Other viable nonlinear optimal control algorithms include iLQR~(\cite{li2004iterative}) and DDP~(\cite{mayne1966second, tassa2014control}).  SAC is particularly well suited for our shared control algorithm because it searches for single, short burst actions which aligns well with our control allocation algorithm (described in detail in Section~\ref{sub-sec-control-allocation}).   Additionally, SAC can compute closed-loop trajectories very quickly (1 kHz), an important feature for interactive human-machine systems such as the one presented in this work. 

\subsubsection{Integrating the Koopman model and SAC}
Sequential Action Control is a gradient-based optimization technique and it is therefore necessary to compute derivatives of a system during the optimization process.  The linearization of the discrete time system is defined by the following equation
\begin{equation*}
x_{t+1} = A  x_t +  B u_t.\\
\end{equation*}

\noindent By selecting a differentiable $\phi$, one can compute $A$ and $B$  
\begin{equation}
\begin{aligned}
A &= \mathcal{K}_{1:N}^T \frac{\partial \phi}{\partial x}\\
B &= \mathcal{K}_{N:N+P}^T \frac{\partial \phi}{\partial u}
\label{eqn-linear-model-a_and-b}
\end{aligned}
\end{equation}

\noindent where $N$ is again the dimension of the state space, and $P$ is the dimension of the control space.  

\subsection{Control Allocation Method}
\label{sub-sec-control-allocation}

To close the loop on our shared control paradigm, we define a control allocation method that uses the solution from the optimal control algorithm to provide outer-loop stabilization.  We use a geometric signal filter that is capable of dynamically shifting which partner is in control at any given instant based on optimality criteria.  This technique is known as Maxwell's Demon Algorithm (MDA)~(\cite{tzorakoleftherakis2015controllers}).  Our specific implementation of MDA is detailed in Algorithm~\ref{mda-algorithm} where $u_h$ is the control input from the human operator, $u_a$ is the control produced by the autonomy, and $u$ is applied to the dynamic system.  
\begin{algorithm}[!h]
\caption{Maxwell's Demon Algorithm (MDA)}
\begin{algorithmic}
  \If {$\langle u_h, u_a \rangle \geq 0$} 
   \State $u = u_h$;
  \Else
   \State $u = 0$;
  \EndIf
\end{algorithmic}
\label{mda-algorithm}
\end{algorithm}

\begin{figure}[!h]
	\centering
	\includegraphics[width=\hsize]{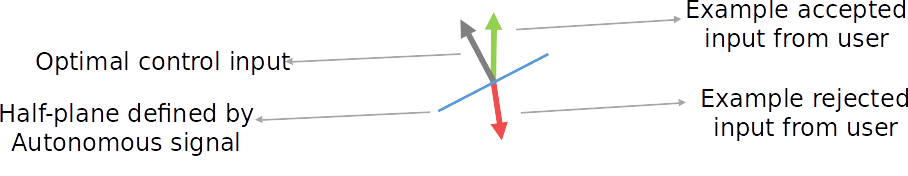}
	\caption{Maxwell's Demon Algorithm (MDA)}
	\label{fig-mda}
\end{figure}

We also provide a pictorial representation of the algorithm in Figure~\ref{fig-mda}.

This control allocation method restricts the user's input to the system to be in the same half-plane as the optimal control solution, and places no other limitations on the human-machine interaction.  If the user's input is in the opposite half-plane, no input is provided to the system.  This control allocation method is lenient to the human partner, as notably, \textit{the autonomous agent does not add any information into the system} and instead only blocks particularly bad input from the user.  Therefore, \textit{any signal sent to the system originates from the human partner}.  We use this filter because we are motivated by assistive robotics, in which prior research has shown that there is no consensus across users on desired assistance level~(\cite{erdogan2017effect}).  By allowing the user a high level of control freedom, the system encourages input from the human operator and restricts how much authority is granted to the autonomous partner.  This method is depicted in Figure~\ref{fig-pipeline}\textcolor{red}{(c)}. We use MDA in this work primarily because it has been experimentally validated in prior studies on human-machine systems for assistive robotics~(\cite{fitzsimons2016optimal}).  Notably, this method does not guarantee optimal (or even "good") performance, as a human operator could theoretically always provide input orthogonal to the autonomous solution, resulting in no control ever being applied to the system.  However, this technique does allocate a large amount of control authority to the human-in-the-loop, a desirable feature in our motivating application domains.  There are also alternative methods that can be used for similar purposes, including extensions to MDA that incorporate additional information from the autonomous partner, which can be used to improve performance or safety~(\cite{broad2018operation}).  A review paper of alternative control allocation techniques can be found in~\cite{losey2018review}.

\section{Human Subjects Study}
\label{sec-experimental-validation}

Here, we detail the experimental setup that we use to study three main aspects of the described system.  
\begin{itemize}
\item First, our aim is to evaluate the efficacy of model-based shared control as it relates to task success and control skill.  Concurrently, we aim to evaluate the generalizability of the learned system models with respect to a wide range of human operators.
\item Second, we aim to evaluate the efficacy of model-based shared control under an online learning paradigm---specifically, the sample-efficiency of the Koopman operator representation.
\item Finally, we aim to evaluate the impact of nonlinear modeling and policy generation techniques through a comparison to a second human-subjects study that enforces linear constraints on our model-based shared control algorithm.
\end{itemize}

\subsection{Experimental Environment}
\label{sec-experimental-environment}

\begin{figure}[!h]
	\centering
	\fbox{\includegraphics[width=0.85\hsize]{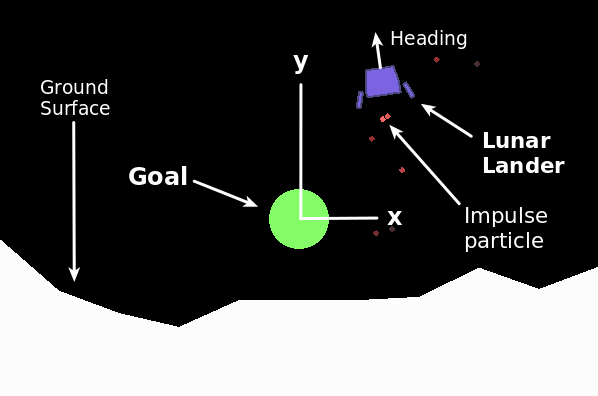}}
	\caption{Simulated lunar lander system.  The green circle is the goal location.  The red dots represent an engine firing.}
	\label{fig-ll-env}
\end{figure}

The proposed shared control framework is evaluated using a simulated lunar lander (see Figure~\ref{fig-ll-env}).  

\noindent We use a simulated lunar lander (rocket) as our experimental environment for a number of reasons.  This environment is challenging for a novice user, but performance can be improved (and sometimes mastered) given enough time and experience.  Similar to a real rocket, one of the main control challenges is the stability of the system.  As the rocket rotates along its yaw axis, firing the main thruster can produce nonintuitive dynamics for a novice.  Furthermore, once the rocket has begun to rotate, momentum can easily overwhelm a user who is unfamiliar with such systems.  Therefore, it is often imperative---particularly for non-expert users---to maintain a high degree of stability at all times in order to successfully complete the task.  In addition to the control challenges, we choose this environment because the simulator abstracts the system dynamics through calls to the Box2D physics engine; therefore, we do not have an exact model and thus have an \textit{explicit need to learn one}.

\subsection{System Description}
\label{sub-sec-system-description}

The dynamic system is a modified version of an open-source environment implemented in the Box2D physics engine and released by OpenAI~(\cite{brockman2016gym}).  Our modifications (1) allow for continuous-valued multi-dimensional user control via a joystick, and (2) incorporate the codebase into the open-source ROS framework.  We have made our code available online at \url{https://github.com/asbroad/model_based_shared_control}.

The lunar lander is defined by a 6D state space made up of the position ($x,y$), heading ($\theta$), and their rates of change ($\dot{x}, \dot{y}, \dot{\theta}$).  The control input to the system is a continuous two dimensional vector ($u_1, u_2$) which represents the throttle of the main and rotational thrusters.  The main engine can only apply positive force.  The left engine fires when the second input is negative, while the right engine fires when the second input is positive.  The main engine applies an impulse that acts on the center of mass of the lunar lander, while the left and right engines apply impulses that act on either side of the rocket.  We remind the reader that our goal is to learn both the system dynamics and user interaction.  For this reason, we must collect data both on the system state and also the control input.  Together, this defines an eight dimensional system: 
\begin{equation*}
\mathcal{X} = [x, y, \theta, \dot{x}, \dot{y}, \dot{\theta}, u_1, u_2]
\end{equation*}

\noindent where the first six terms define the lunar lander state and $u_1, u_2$ are the main and rotational thruster values, through which the user interacts with the system.

\subsection{Trial Description}
\label{sec-experiment-trial-description}

The task in this environment requires the user to navigate the lander from its initial location to the goal location (represented by the green circle in Figure~\ref{fig-ll-env}) and to arrive with a heading nearly perpendicular to the ground plane and with linear and rotational velocities near zero.  A trial is considered complete either (1) when the center of an upright lunar lander is fully contained within the goal circle (i.e., the Euclidean distance between the center of the lander and the center of the goal is less than $0.9$ m) and the linear and angular velocities are near zero (i.e., the linear velocities must be less than $1.0$ m/s and the angular velocity must be less than $0.3$ rad/sec), or (2) when the lander moves outside the bounded environment (i.e., when the lander moves off the screen to the left or right) or crashes into the ground.

\begin{figure}[!h]
	\centering
	\includegraphics[width=0.85\hsize]{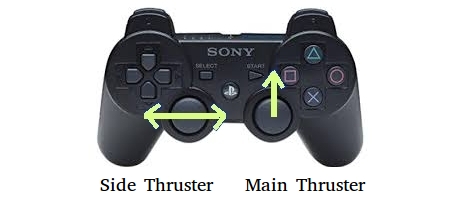}
\end{figure}
In each trial, the lunar lander is initialized to the same $x, y$ position ($10.0$ m, $13.3$ m), to which we added a small amount of Gaussian noise ($\mu = 0.2$ m).  Additionally, a random force is applied at the start of each trial (uniform($-1000$ N,$1000$ N)).  The goal location ($10.0$ m, $6.0$ m) is constant throughout all trials and is displayed to the user as a green circle (see Figure~\ref{fig-ll-env}).  

The operator uses a PS3 controller to interact with the system.  The joystick controlled by the participant's dominant hand fires the main thruster, and the opposing joystick fires the side thrusters.  As the user moves through the environment, we keep track of the full state space at each timestep (10 Hz).  We provide a video of the system, task and user interaction under shared control as part of the supplementary material.

\subsection{Analysis I : Efficacy and Generalizability of Model-based Shared Control}
\label{sub-sec-analysis-i}

\subsubsection{Control Conditions}

To study the efficacy and generalizability of our shared control system and the generalizability of the learned system dynamics, we compare four distinct control conditions.

\begin{itemize}
\item In the first condition, the user is in full control of the lander and is not assisted by the autonomy in any way; we call this approach \textit{User Only} control.  As each user undergoes repeated trials with the same goal, this can also be considered a natural learning paradigm.  
\end{itemize}

In the remaining three conditions an autonomous agent provides outer-loop stabilization on the user's input as described in Section~\ref{sec-model-based-shared-control}.  The main distinction between these three control conditions is the source of the data used to compute the model of the joint system.  
\begin{itemize}
\item In the second condition, the model is defined by a Koopman operator learned on data captured from earlier observations of the current user; we call this approach \textit{Individual Koopman}.  
\item In the third condition, the model is defined by a Koopman operator learned on data captured from observations of three novice participants prior to the experiment (who were not included in our analysis); we call this approach \textit{General Koopman}.
\item In the fourth condition, the model is defined by a Koopman operator learned on data captured from observations of an expert user (the first author of the paper, who has significant practice controlling the simulated system); we call this approach \textit{Expert Koopman}.
\end{itemize}

We analyze the viability of model-based shared control by comparing the \textit{User Only} condition to each of the shared control conditions.  We analyze the generalizability of the learned models by comparing the results from the \textit{Individual Koopman}, \textit{General Koopman} and \textit{Expert Koopman} conditions.  All of the data we analyze to evaluate the ideas presented in this section comes from the nonlinear MbSC study.

\subsubsection{Protocol and Participants}

Each experiment begins with a training period for the user to become accustomed to the dynamics of the system and the interface. This training period continues until the user is able to successfully achieve the task three times in a row or 15 minutes elapses.  During the next phase of the experiment, we collect data from 10 user-controlled trials, which we then use to develop a model.  Finally, each user performs the task under the four conditions detailed above (10 trials each).  The order in which the control paradigms are presented to the user is randomized and counter-balanced to reduce the effect of experience. 

The study consisted of 16 participants (11 female, 5 male).  All subjects gave their informed consent and the experiment was approved by Northwestern University's Institutional Review Board. 

\subsection{Analysis II : Online Model-based Shared Control}
\label{sub-sec-analysis-ii}

To study the efficacy of our model-based shared control algorithm in an online learning paradigm, we collect data from a fifth experimental condition, which we call \textit{Online Koopman}.  

\subsubsection{Control Condition}

\begin{itemize}
\item The main difference between the \textit{Online Koopman} paradigm and the three previously described shared control conditions is that the model of the joint human-machine system is learned online in real-time.  In all other control conditions, all models were trained offline from observations gathered during a data collection phase.  In the online paradigm, the model is updated continuously starting with the first collected set of observations.
\end{itemize}

In addition to the lack of a separate data collection phase, the online learning paradigm is distinct from the other shared control conditions because of the data that we use to learn the model.  In the shared control conditions that use a model learned offline, we use all of the observations collected from the user demonstrations to learn the model.  In the online learning paradigm we only update the model when the user input is admitted by the MDA controller.  We choose this learning paradigm because it fits well conceptually with our long term goal of using the outer-loop algorithm to provide stability and safety constraints on the shared control system.  It is important to note that at the beginning of the online learning paradigm, the MDA controller relies on randomly initialized control and system dynamics models.  For this reason, the control let through to the system will be very noisy during the first few moments of the experiment, making the system difficult to control successfully for any human-in-the-loop.  For this reason, it is important that the system dynamics and control models can be learned quickly, something we evaluate in Section~\ref{sec-study-two-results-online}.  As soon as the learning process produces a model of any kind, the policy is computed using MPC techniques.

\subsubsection{Protocol and Participants}
\label{sec-sub-sub-protocol}

The online learning paradigm consists of 15 trials per user to allow us to evaluate possible learning curves.  The model is updated at the same rate as the simulator (10 Hz) and is initialized naively (i.e., all values are sampled from a uniform distribution [0,1)).  This paradigm is presented as the final experimental condition to all subjects.  The subjects are the same 16 participants as in Section~\ref{sub-sec-analysis-i}. All of the data we analyze to evaluate the ideas presented in this section comes from the nonlinear MbSC study.

\subsection{Analysis III : Comparison of Linear and Nonlinear Model-based Shared Control}
\label{sub-sec-analysis-iii}

To study the impact of nonlinear modeling and policy generation techniques on our model-based shared control paradigm, we compare results from the above study to a second study (consisting of a separate group of 16 participants) that enforces linear constraints on these parts of the system.  

\subsubsection{Control Conditions}

The same four control conditions from Analysis I are evaluated.  The differences lie in (1) the choice of basis function used to approximate the Koopman operator and (2) the choice of optimal control algorithm used to generate the autonomous policy.  In this study, we use a linear basis, instead of a nonlinear basis, to approximate the Koopman operator, which consists of the first nine terms in Equation~\eqref{eq-basis}.  We furthermore use a Linear Quadratic Regulator, instead of a nonlinear model predictive control (MPC) algorithm (Sequential Action Control (SAC)).  

\subsubsection{Protocol and Participants}

The same experimental protocol described in Section~\ref{sub-sec-analysis-i} was used, allowing us to perform a direct comparison between the two studies.  Unlike the prior sections, the data we analyze to evaluate the ideas presented in this section comes from both the linear and nonlinear MbSC studies.  The data from the linear MbSC study was previously analyzed in~(\cite{broad2017learning}) and was collected from a different set of 16 subjects, resulting in 32 total participants.

\subsection{Statistical Analysis} 
\label{sec-statistical-analysis}

We analyze the results of the human-subjects studies using statistical tests to compare the performance of participants along a set of pertinent metrics under the control conditions described in Section~\ref{sub-sec-analysis-i}.  Our analysis consists of one-way ANOVA tests conducted to evaluate the effect of the shared control paradigm on each of the dependent variables in the study.  These tests allow us to statistically analyze the effect of each condition while controlling for overinflated type I errors that are common with repeated t-tests.  Each test is computed at a significance value of 0.05.  When the omnibus F-test produces significant results, we conduct post-hoc pair-wise Student's t-tests using Holm-Bonferroni adjusted alpha values~(\cite{wright1992adjusted}).  The post-hoc t-tests allow us to further evaluate the cause of the significance demonstrated by the ANOVA by comparing each pair of control paradigms separately.  Similar to the ANOVA test, the Holm-Bonferroni correction is used to reduce the likelihood of type I errors in the post-hoc t-tests.  

In addition to reporting the results of the statistical tests, we also use box-and-whisker diagrams to display specific metrics.  In these plots, the box represents the \textit{interquartile range (IQR)} which refers to the data that lies between the first and third quartiles.  This area contains 50$\%$ of the data.  The line inside the box represents the median value and the whiskers above and below the box are the minimum and maximum values inside 1.5 times the interquartile range.  The small circles are outliers.  The plots also depict the results of the reported statistical tests.  That is, if a post-hoc t-test finds statistically significant differences between two conditions, we depict these results on the box-and-whisker diagrams using asterisks to represent the significance level ($*: p < 0.05, **: p < 0.01, ***: p < 0.005$). 

We note that this analysis is used for \textit{all reported results}.  Therefore, if we present the results of a t-test, it signifies that we have previously run an ANOVA and found a statistically significant difference.  The reader can also assume that any unreported post-hoc t-tests failed to reject the null hypothesis.  

\section{Results}
\label{sec-results}

We now present the results of the desired analyses described in Sections~\ref{sub-sec-analysis-i}, ~\ref{sub-sec-analysis-ii}, and ~\ref{sub-sec-analysis-iii}.  Our analyses support the premise that model-based shared control is a valid and effective data-driven method for improving a human operator's control of an \textit{a priori} unknown dynamic system.  We also find the learned system models are generalizable across a population of users.  Finally, we find that these models can be learned online in a fast, data-efficient manner.

\subsection{Efficacy of Model-based Shared Control}
\label{sec-results-performance-metrics}

To evaluate the efficacy of our model-based shared control algorithm, we compute the average success rate under each control paradigm and examine the distribution of executed trajectories.  Our analysis compares the User Only control condition to each of the shared control conditions (Individual Koopman, General Koopman and Expert Koopman). All of the data we analyze to evaluate the ideas presented in this section comes from the nonlinear MbSC study.

\subsubsection{Task Success and User Skill}

A trial is considered a success when the user is able to meet the conditions defined in Section~\ref{sec-experiment-trial-description}.  We can interpret the success rate of a user, or shared control system, on a set of trials as a measure of skill.  The greater the skill, the higher the success rate.  By comparing the average success rate under the User Only control paradigm with the average success rate under the shared control paradigms, we can analyze the impact of the assistance provided by the autonomous partner.  

\begin{figure}[h]
  \centering
  \includegraphics[width=0.8\hsize]{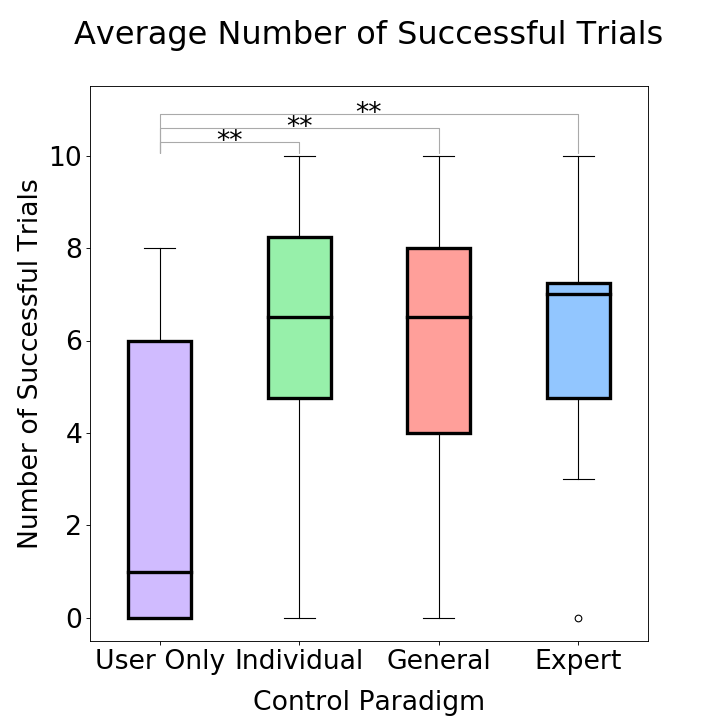} 
  \caption{Number of successful trials under each control condition.}
  \label{fig-results-success-nonlinear}
\end{figure}  

\begin{figure*}[t]
	\centering
	\includegraphics[width=0.8\hsize]{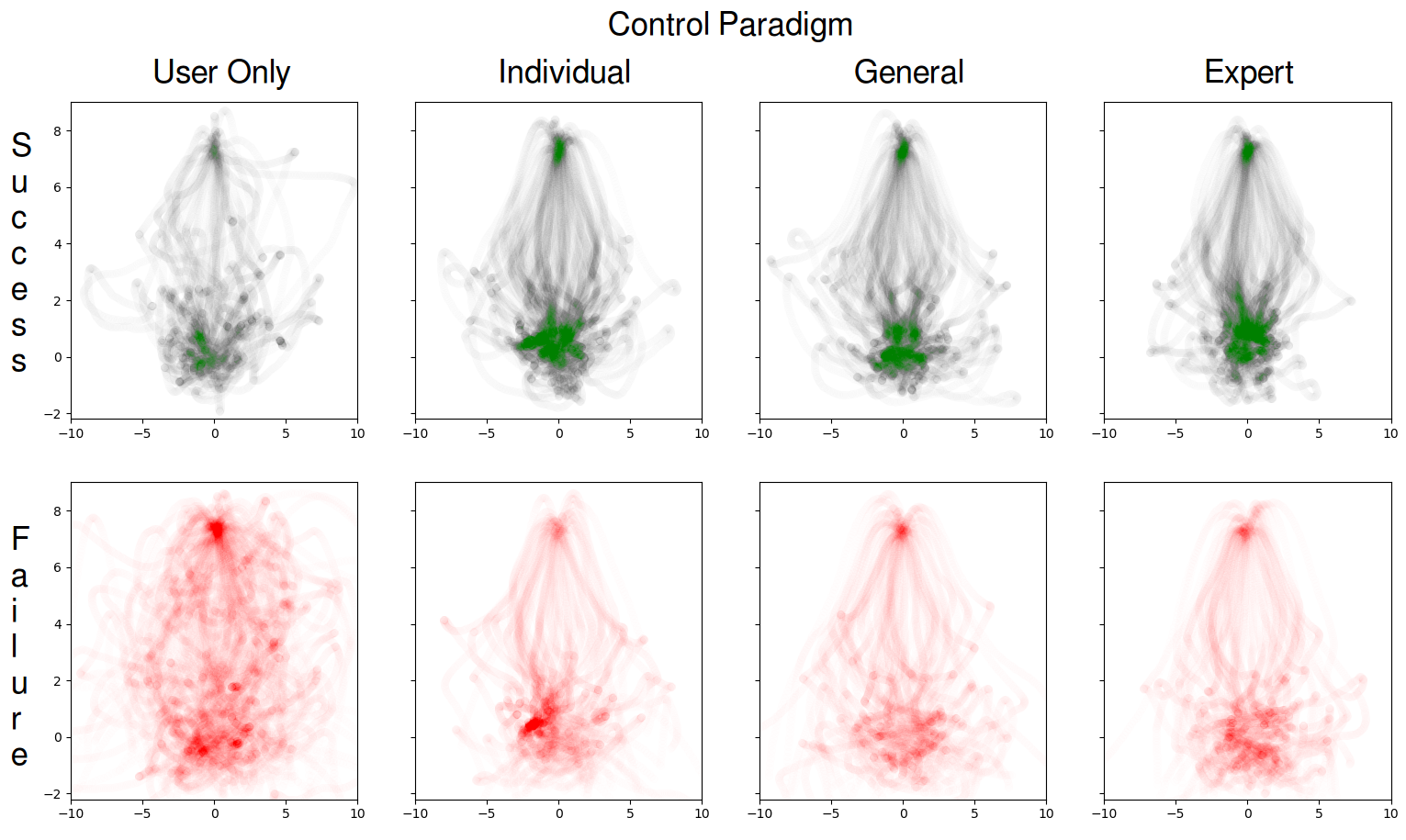}
	\caption{Trajectory plots which visualize the most frequently visited parts of the state space.  The data is broken down by control condition (columns) and whether the trial was successful (rows).  The plots are generated by overlaying each trajectory with a low opacity and the intensity of the plots therefore represents more frequently visited portions of the state space.}
	\label{fig-heatmaps}
\end{figure*}

The average number of successful trials produced in each control condition are displayed in Figure~\ref{fig-results-success-nonlinear}. An analysis of variance shows that the choice of control paradigm has a significant effect on the success rate ($F(3, 59) = 4.58, p < 0.01$). Post-hoc t-tests find that users under the shared control conditions show statistically significant improvements in the success rate when compared to their performance under the User Only control condition ($p < 0.01$, for all cases).  No other pairings are found to be statistically distinct.  This result demonstrates that the assistance provided by the autonomous agent significantly improves the skill of the joint human-machine system, thereby validating the efficacy of model-based shared control.  This result is inline with related work that aims to provide similar assistance, such as can be found in the virtual fixtures/guides literature~(\cite{forsyth2005predictive, griffiths2005sharing, abbink2012haptic}).  Importantly, however, unlike these prior methods, model-based shared control does not require \textit{a priori} knowledge of the system dynamics or the human operator.

The fact that there are no observed differences in task performance between the Individual, General and Expert cases suggests that the source of the data used to learn the model may not be important in developing helpful autonomous assistance in the shared control of dynamic systems (discussed further in Section~\ref{sec-results-generalizability}).  An alternative interpretation of this data could be that the discrepancy of skill demonstrated by the participants in the individual, general and expert cases was not large enough to produce any potential difference in performance.  This, however, is not likely, as the expert demonstrator (the lead author) was able to easily achieve the desired goal state during every demonstration (i.e. 10 out of 10 trials).  In contrast, the average subject who provided data in the individual and general cases performed similarly to how participants performed under the User Only cases in Figure~\ref{fig-results-success-nonlinear} (i.e. about 1 in 10 successful demonstrations).

\subsubsection{Distribution of Trajectories---Qualitative}

We further analyze the different control conditions through a comparison of the distribution of trajectories we observe in each condition.  Unlike the success metric, this analysis is not based on task performance, and is instead performed to evaluate the control skill exhibited by either the human operator alone or the joint human-machine system.  Figure~\ref{fig-heatmaps} depicts trajectory plots which represent the most frequently occupied sections of the state space.  The plots are generated using data separated based on the control condition (columns) and whether the user was able to complete the task on a given trial (rows). 

The first distinction we draw is between the User Only control condition and the three shared control conditions.  In particular, the distribution of trajectories in the User Only condition depicts both larger excursions away from the target and lower levels of similarity between individual executions.  When we focus specifically on which parts of the state space users spend the most time in (as represented by the intensity of the plots), we see two main clusters of high intensity (around the start and goal locations) in the shared control conditions, whereas we see a wider spread of high-intensity values in the User Only control condition.  This suggests more purposeful motions under the shared control conditions.

The second distinction we draw focuses on a comparison between the successful and unsuccessful trials.  Specifically, we note that trajectory plots computed from the failed trials under the shared control conditions demonstrate similar properties (e.g., the extent of the excursions away from the target, as well as two main clusters of intensity) to the trajectory plots computed from successful trials under the shared control conditions.  This suggests that users may have been closer to succeeding in these tasks than the binary success metric gives them credit for.  By comparison, the trajectory plot computed from the failed trials under the User Only control condition depicts a significantly different distribution of trajectories with less structure.  Specifically, we observe numerous clusters of intensity that represent time spent far away from the start and goal locations.  This suggests that users were particularly struggling to control the system in these cases.

\subsubsection{Distribution of Trajectories---Quantitative}

These observations are supported by an evaluation of the ergodicity~(\cite{mathew2011metrics}) of the distributions of trajectories described above.  We find users under the shared control paradigm are able to produce trajectories that are more ergodic with respect to the goal location then users under User Only control, which means that they spend more a significantly larger proportion of their time navigating near the goal location under shared control.

To perform this comparison, we compute the ergodicity of each trajectory with respect to a probability distribution defined by a Gaussian centered at the goal location (which represents highly desirable states).  This metric can be calculated as the weighted Euclidean distance between the Fourier coefficients of the spatial distribution and the trajectory~(\cite{miller2016ergodic}).

Similar to our qualitative analysis of the trajectory plots in Figure~\ref{fig-heatmaps}, we first compare ergodicity between the different control conditions by analyzing \textit{all} the trajectories observed under each condition.  An analysis of variance showed that the effect of the shared control paradigm on trajectory ergodicity is significant ($F(3, 640) = 12.97, p < 0.00001$).  Post-hoc t-tests find statistically significant differences between the performance of the users in the User Only control condition and users in the shared control conditions based on the individual, general and expert datasets ($p < 0.0005, p < 0.001, p < 0.0005$, respectively).  No other pairings demonstrate statistically distinct results. We interpret this result as additional evidence that model-based shared control improves the skill of the human partner in controlling the dynamic system.

We further analyze the ergodicity results by separating the trajectories based on whether they come from an unsuccessful or successful trial.  An analysis of variance computed over all control conditions showed that the effect of the shared control paradigm on trajectory ergodicity is significant for both unsuccessful ($F(3, 310) = 6.60, p < 0.0005$) and successful ($F(3, 325) = 7.20, p < 0.0005$) trials.  Post-hoc t-tests find statistically significant differences between the performance of the users in the User Only control condition and users in the shared control conditions ($p < 0.005$ in all unsuccessful cases, $p < 0.05$ in all successful cases).  No other pairings reject the null hypothesis.  These results suggest that the shared control paradigm is helpful in improving the user's skills even when they provide input that is ultimately unsuccessful in achieving the task.  Furthermore, our shared control paradigm is helpful, even when users are performing at their best.  Thus, for both failed and successful trials, users exhibit a greater amount of control skill than when there is no assistance. 

\subsection{Generalizability of Shared Control Paradigm}
\label{sec-results-generalizability}

We continue the evaluation of our human subjects study with an analysis of the generalizability of the learned system models and our model-based shared control algorithm.  All of the data we analyze to evaluate the ideas presented in this section comes from the nonlinear MbSC study.  As reported in Section~\ref{sec-results-performance-metrics}, we find no statistical evidence that the source of the data used to train the model impacts the efficacy of the shared control paradigm.  This test was again conducted using an ANOVA which can be used to evaluate differences between groups by comparing the mean and variance computed from the data collected during the experimental trials. When we compare the success rate of users in each shared control condition, we find no statistically significant difference.  However, we do find a significant difference between the user's performance under each shared control condition and the User Only condition.  The same result holds when we compare each control condition along the ergodic metric described in Section ~\ref{sec-results-performance-metrics} and visualized by trajectory plots in Figure~\ref{fig-heatmaps}.  Taken together, these results suggest that the efficacy of the assistance provided by the autonomous agent is \textit{independent of the source of the data used to learn a model of the joint system}.  That is, models trained on data collected from an individual user generalize to a larger population of human partners. 



\begin{figure}[h]
  \centering
  \captionsetup[subfigure]{justification=centering}
  \begin{subfigure}[t]{0.49\hsize}
  	\centering
  	\includegraphics[width=\hsize]{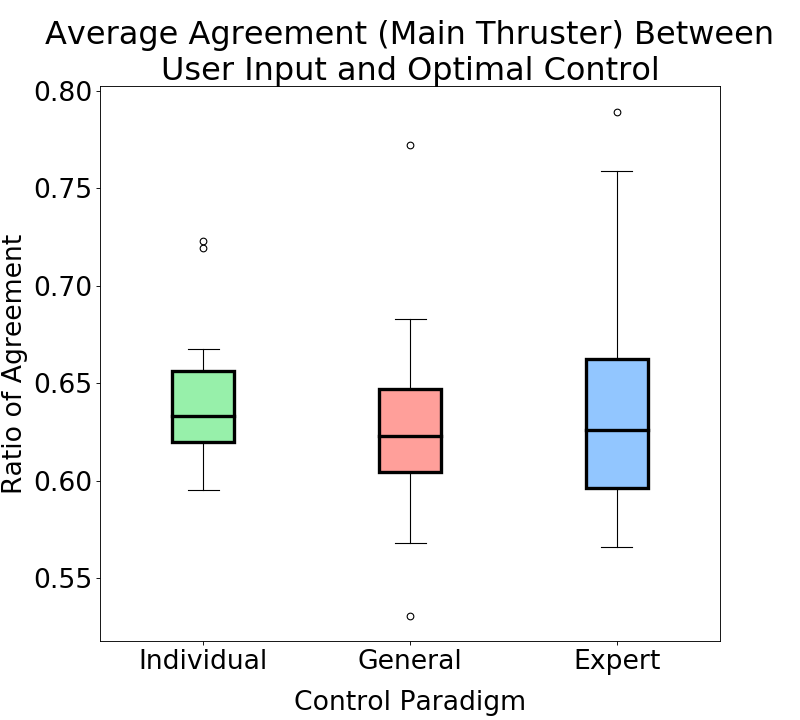} 
  	\caption{}
  	\label{fig-results-agreement-main}
  \end{subfigure}
  \begin{subfigure}[t]{0.49\hsize}
    \centering
    \includegraphics[width=\hsize]{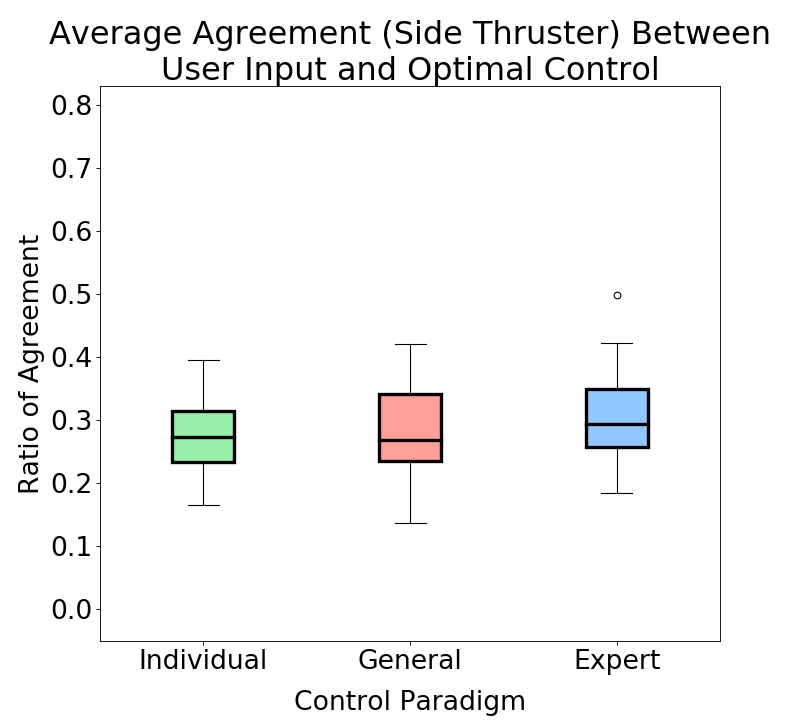} 
    \caption{}
    \label{fig-results-agreement-side}
  \end{subfigure}
  \caption{Average agreement between user and optimal control algorithm as defined by the Maxwell's Demon Algorithm (Equation~\eqref{mda-algorithm}) along the (\subref{fig-results-agreement-main}) main and (\subref{fig-results-agreement-side}) side thrusters.}
  \label{fig-results-agreement}
\end{figure}  

To further analyze the generalizability of the model-based shared control paradigm, we evaluate the participants' interactions with the outer-loop autonomous control.  We are interested in whether or not users agree more often with the autonomy when control signals are produced based on models learned from their personal demonstration data.  To evaluate this idea, we look at the percentage of user inputs that are let through as control to the dynamic system based on our control allocation method (MDA).  The average agreement metric is broken down by control condition and presented in Figure~\ref{fig-results-agreement}.  

An analysis of variance shows that the effect of the source of the model data on the average agreement is not significant in either the main thruster ($F(2, 44) = 0.87, p = 0.43$) nor the side thruster ($F(2, 44) = 0.38, p = 0.69$).  These results show a uniformity in the response to system state across users and suggest that the system is able to adapt to the person, instead of requiring a personalized notion of the user and system.  

We interpret this finding as further evidence of the generalizability of our model-based shared control paradigm.  In particular, we find that it is not necessary to incorporate demonstration data from individual users when developing model-based shared control.  This result replicates findings from our analysis of data collected under a shared control paradigm that enforced a linear constraint on the model learning and policy generation techniques~(\cite{broad2017learning}).

\subsection{Online Learning Shared Control}
\label{sec-study-two-results-online}

\begin{figure}[!h]
    \centering
    \includegraphics[width=\hsize]{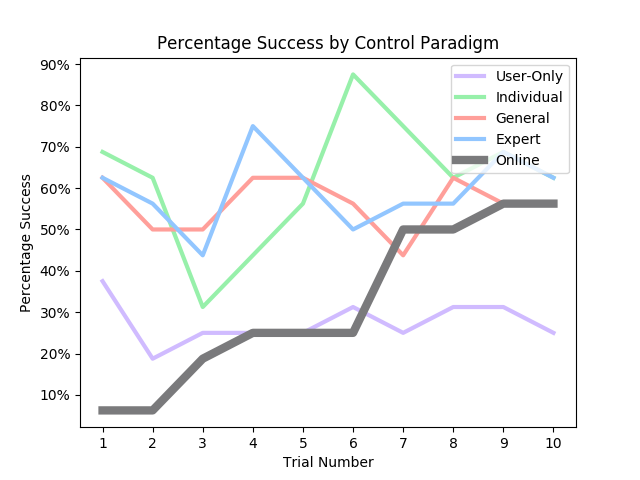} 
    \caption{Average percentage success by trial for the first 10 trials by control condition.  Users under all shared control conditions using models learned offline (Individual, General, Expert) outperform the User Only control condition across all trials.  Users under the shared control condition using models learned online (Online) start off performing poorly, but quickly begin to outperform the User Only control condition and, in the end, achieve the same level of success as those under the offline shared control conditions.}
    \label{fig-res-success-by-trial}
\end{figure}

We next evaluate our model-based shared control algorithm in an online learning paradigm.  Our evaluation considers the sample complexity of our model-based learning algorithm through a comparison of the \textit{impact each shared control paradigm has on the skill of the joint system over time}.  All of the data we analyze to evaluate the ideas presented in this section comes from the nonlinear MbSC study.  Our statistical analysis is a comparison of the percent of participants who succeed under each paradigm \textit{by trial number}, shown in Figure~\ref{fig-res-success-by-trial}.  We remind the reader that users participate in 15 trials of the Online Koopman condition while they participate in 10 trials of the four other experimental conditions.  For comparison we only plot the first 10 trials of the Online Koopman data, though we note that the improved success rate is sustained over the final five trials.  From this plot, we can see that users in the Online Koopman shared control condition start off performing poorly, but by around trial 7 start performing on par with the other shared control conditions.  

Here, we also note the number of trials used to train the model of the system and control dynamics in each condition.  In the Individual and Expert conditions, data is collected from 10 trials to train the model.  In the General condition, data is collected from three different users who each control the system for 10 trials each, which means the model is trained from a total of 30 trials.  Finally, as discussed above, in the Online condition, the model is learned continuously over the course of 15 trials.

To provide quantitative evidence of this visual trend, we perform the same types of statistical analyses as in previous sections, but now include data from the \textit{Online Koopman} as a fifth experimental condition.  For ease of discussion we refer to the \textit{Individual}, \textit{General} and \textit{Expert Koopman} model-based shared control conditions as the offline learning conditions, and the \textit{Online Koopman} model-based shared control as the online learning condition.  As users provide more data in the Online Koopman condition than in all other conditions, we perform two sets of analyses.  First, we compare the data from the first ten trials from the Online Koopman condition to all other control conditions.  We then re-perform the same tests, but use the final ten trials from the Online Koopman condition.  By comparing these results, we can evaluate the efficacy of the online learning paradigm, and also analyze the effect of the amount of data used during the learning process. 

\subsubsection{Statistical Analysis of the First Ten Trials}
\label{sec-study-two-results-online-first-ten}

An analysis of variance finds a statistically significant difference between the various control conditions along the primary success metric ($F(4, 74) = 5.35, p < 0.001$).  Post-hoc t-tests find that all offline learning conditions significantly outperform the Online Koopman and User Only control conditions ($p < 0.05$ for all cases).  We do not find the same statistically significant difference between the User Only and Online Koopman conditions.  These results suggest that users under the online learning paradigm initially perform on par with how they perform under the user only control paradigm, but worse than under the offline control conditions.  This analysis is consistent with our expectations since, in the online condition, the model of the joint system is initialized randomly and therefore does a poor job of assisting the user.  However, it is also important that this online shared control does not degrade performance in comparison to the User Only paradigm, suggesting that there is little downside to employing the online learning paradigm during learning.  

\subsubsection{Statistical Analysis of the Final Ten Trials}
\label{sec-study-two-results-online-last-ten}

\begin{figure}[!h]
    \centering
    \includegraphics[width=0.8\hsize]{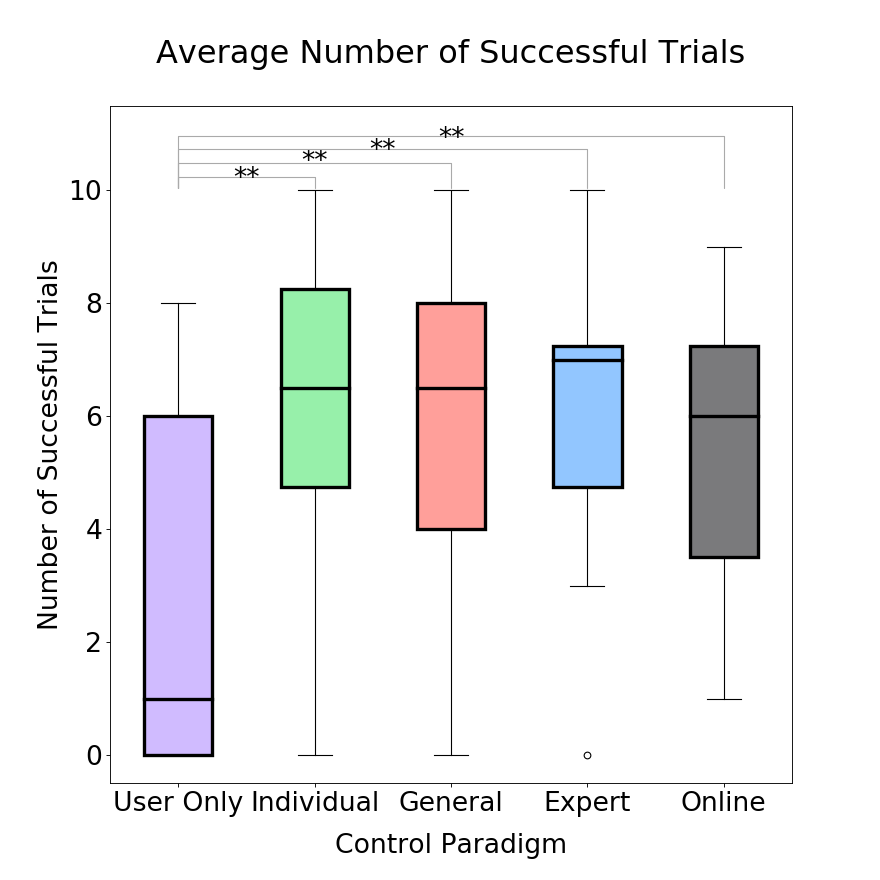} 
    \caption{Number of successful trials under each control condition (including an online learning paradigm).  We find statistically significant differences between the User Only condition and each shared control condition ($p < 0.01$).}
    \label{fig-avg-metrics-second-online-study}
\end{figure}

As a point of comparison, we now re-run the same statistical tests using the final ten trials from the Online Koopman condition.  An analysis of variance finds a statistically significant difference between the various control conditions along the primary success metric ($F(4, 74) = 3.55, p < 0.05$) (see Figure~\ref{fig-avg-metrics-second-online-study}).  Post-hoc t-tests find that all shared control conditions (using models learned offline and online) significantly out perform the User Only control paradigm ($p < 0.01$ for all conditions).  This result is different from our analysis of the first ten trials and suggests that the learned model improves significantly with more data and now is on par with the models learned in the offline conditions.  No other pairings show statistically significant differences.  

The visual trend present in Figure~\ref{fig-res-success-by-trial} and the statistical analysis demonstrated in Figure~\ref{fig-avg-metrics-second-online-study} suggest that the Koooman operator is able to quickly learn an actionable representation of the joint human-machine system.  These results also demonstrate the efficacy of our model-based shared control algorithm in an online learning scenario and in limited data regimes.  Here we note that follow-up studies are required to tease apart the impact of the model learning process and the user's experience controlling the dynamic system when comparing the offline paradigms to the online paradigm.  Notably, in the offline learning paradigm, users undergo 10 trials of training at the start of the experiment.  In contrast, and as stated in Section~\ref{sec-sub-sub-protocol}, all users operate the system under the online learning paradigm as the final condition.  For this reason, we do not account for user experience in this condition and therefore highlight the data-efficiency of the model learning process instead of the overall task performance in this condition.  The main takeaway from this portion of the analysis is therefore that an actionable Koopman operator can be learned \textit{quickly}, from significantly less data than alternative approaches commonly found in the literature, like neural networks~(\cite{nagabandi2018neural}).

\subsection{Linear and Nonlinear Model-based Shared Control}

The final piece of analysis we perform in this work is related to the impact that nonlinear modeling and policy generation techniques have on our model-based shared control paradigm.  For this analysis we compare the User Only control condition to the three offline shared control conditions.  Unlike the prior sections, the data we analyze to evaluate the ideas presented in this section comes from both the linear and nonlinear MbSC studies.

\begin{figure}[h]
  \centering
  \captionsetup[subfigure]{justification=centering}
  \begin{subfigure}[t]{0.49\hsize}
  	\centering
  	\includegraphics[width=\hsize]{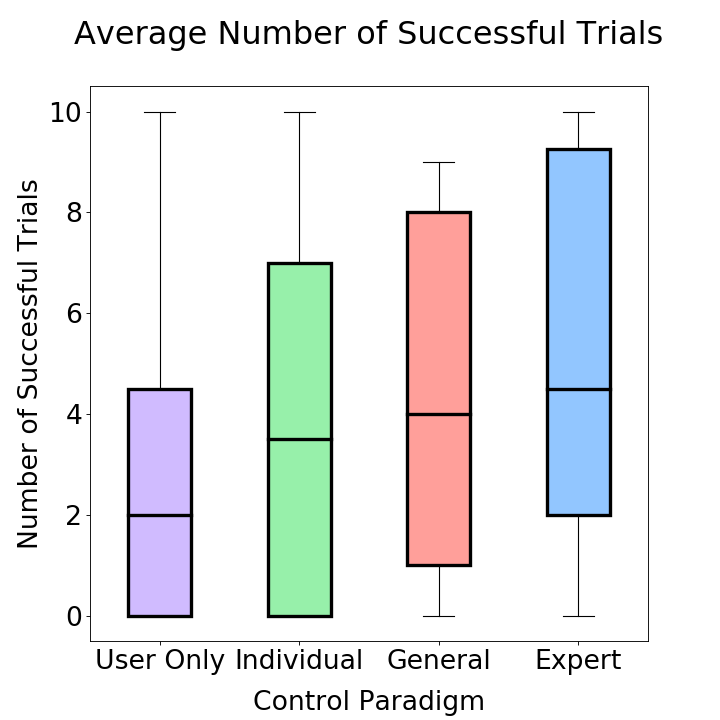} 
  	\caption{Linear MbSC.}
  	\label{fig-results-comparison-success-linear}
  \end{subfigure}
  \begin{subfigure}[t]{0.49\hsize}
  \centering
  \includegraphics[width=\hsize]{results_success_nonlinear.png} 
  \caption{Nonlinear MbSC.}
  \label{fig-results-comparison-success-nonlinear}
  \end{subfigure}
  \caption{A comparison of the average success rate under (a) linear and (b) nonlinear model-based shared control to user only control.}
  \label{fig-results-comparison-success}
\end{figure}  


The average success rate of users under each control paradigm for both studies is presented in Figure~\ref{fig-results-comparison-success}.  In the linear study~(\cite{broad2017learning}) we observe a trend (see Figure~\ref{fig-results-comparison-success-linear}) that suggests users perform better under the shared control paradigm, but we do not find statistically significant evidence of this observation.  In contrast, we find that model-based shared control using nonlinear modeling and policy generation techniques does statistically improve the success rate when compared to a User Only control paradigm.

\begin{figure}[!h]
	\centering
	\includegraphics[width=0.9\hsize]{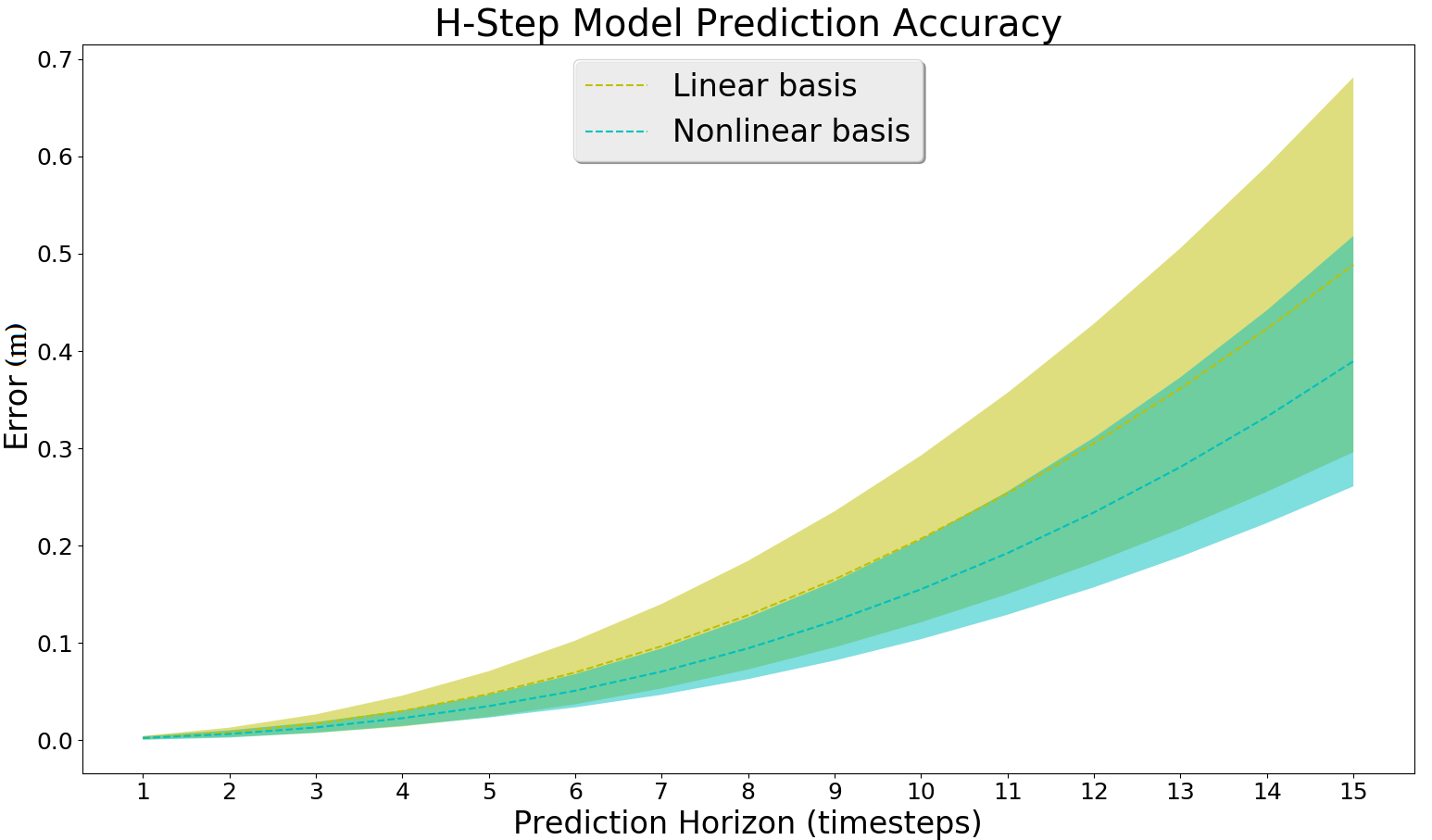}
	\caption{H-step Prediction Accuracy of Koopman operator models based on linear and nonlinear bases.  Error is computed as a combination of the Euclidean distance between the predicted $(x, y)$ values and the ground truth $(x, y)$.}
	\label{fig-h-step-accuracy}
\end{figure}

One potential explanation for the difference we find in the results of the two studies is that the nonlinear basis produces more accurate models of the system dynamics then the linear basis.  To explore this explanation, we evaluate the predictive capabilities of a Koopman operator learned with a linear basis to one learned with a nonlinear basis.  This analysis is performed by comparing the predicted system states with ground truth data.  We evaluate the error (mean and variance) as a function of prediction horizon (a.k.a. the H-step error).  Figure~\ref{fig-h-step-accuracy} depicts the raw error (in meters) of Koopman operators trained using linear and nonlinear bases.  

Our analysis of the predictive capabilities of the Koopman operator models demonstrates that each is highly accurate.  The nonlinear model does slightly outperform the linear model as the prediction horizon grows; however, we find that both models are able to produce single-step predictions with error on the scale of $10^{-3}$ meters.  As a reminder to the reader, the state space is bounded with $X \in (-10, 10), Y \in (0, 16)$.  This analysis suggests that the choice of basis function does not cause the observed difference in average success rate between the two studies.  Instead, the important design decision may be the choice of model predictive control algorithm.  In the linear study we use an infinite horizon LQR to produce autonomous control policies, whereas in the nonlinear study we use a receding-horizon Model Predictive Control (MPC) to produce autonomous control.  Our interpretation of these results is that the receding horizon nature of MPC is better suited to the visual planning approach that human operators use when solving the lunar lander task.

\section{Discussion}
\label{sec-discussion}

In this section, we highlight a number of main takeaways that stem from our analysis.  To begin, the results of our human-subjects studies demonstrate that our model-based shared control paradigm is able to (1) successfully learn a model of the joint human-machine system from observation data, and (2) use the learned system model to generate autonomous policies that can help assist a human partner achieve a desired goal.  We evaluate the predictive capabilities of the learned system models through a comparison to ground truth trajectory data (see Figure~\ref{fig-h-step-accuracy}) and evaluate the impact of the assistive shared control system through a comparison of performance (success rate, see Figure~\ref{fig-results-success-nonlinear}) with a User Only (or natural learning) control paradigm.  All analyses support the idea that MbSC can help improve the control skill of a human operator both when they are able to achieve a task on their own and when they are not.

Additional evaluations demonstrate that the learned system and control dynamics generalize across users, and suggests that, unlike in other human-machine interaction paradigms~(\cite{sadigh2016information, macindoe2012pomcop, javdani2015shared}), personalization is not required for defining shared control paradigms of generic human-machine systems.  Specifically, we find that the demonstration data used to learn the system and control models does not need to come from an optimal, or expert, controller, and can instead come from \textit{any} human operator.  Therefore, at a base level, the controller does not need to be personalized to each individual user as the learned model captures all necessary information.  This idea is important for application in real-world scenarios where personalization of control paradigms can be time-consuming, costly, and challenging to appropriately define, often due to the variety in preferences described by human operators~(\cite{gopinath2016human, erdogan2017effect}).

We also demonstrate that our approach can be used in an online learning setting.  Importantly, we find that the model is able to learn very quickly, from limited amounts of data.  In the Online Koopman condition, each trial took an average of 18 seconds, and therefore provided 180 data points. From our analysis in Section~\ref{sec-study-two-results-online-last-ten}, we find that we are able to learn an effective model of the joint system after only 5 trials (or about 900 data points). Our model learning technique is also well suited for an online learning paradigm as it is not computationally intensive and can easily run at 50Hz on a Core i7 laptop with 8 GB of RAM.  Additionally, we find that, even during the learning process, the application of the online model-based shared control algorithm does not significantly degrade the performance of the human operator.

Finally, we also evaluate the impact that nonlinear modeling and policy generation techniques have on our model-based shared control algorithm~(\cite{broad2017learning}).  In particular, we replace the nonlinear modeling and policy generation techniques with linear counterparts and compare how they impact the ability of a human operator to achieve a desired task.  This requires using a nonlinear basis when computing the approximation to the Koopman operator and using nonlinear model predictive control (SAC) to generate the autonomous policy.  We find that the nonlinear model-based shared control paradigm produces a joint human-machine system that is significantly better along the primary performance metric (task success) then users under a user only control paradigm.  The same result is not found from the data collected under a shared control paradigm that enforced linear constraints (see Figure~\ref{fig-results-comparison-success}).

\section{Conclusion}
\label{sec-conclusion}

In this work, we introduce model-based shared control (MbSC).  A particularly important aspect of this work is that \textit{we do not rely on a priori knowledge, or a high-fidelity model, of the system dynamics}.  Instead, we learn the system dynamics \textit{and} information about the user interaction with the system directly from data.  We learn this model through an approximation to the Koopman operator, an infinite dimensional linear operator that can exactly model non-linear dynamics.  By learning the joint system dynamics through user interaction, the robot's understanding of the human is implicit to the system definition.

Results from two human subjects studies (consisting of 32 total participants) demonstrate that incorporating the learned models into our shared control framework statistically improves the performance of the operator along a number of pertinent metrics.  Furthermore, an analysis of trajectory ergodicity demonstrates that our shared control framework is able encourage the human-machine system to spend a significantly greater percentage of time in desirable states.  We also find that the learned system models are able to be used in shared control systems that generalize across a population of users. Finally, we find that, using this approach, models can be efficiently learned online.  In conclusion, we believe that our approach is an effective step towards shared control of human-machine systems with unknown dynamics.  This framework is sufficiently general that it could be applied to any robotic system with a human in the loop.  Additionally, we have made our code available online at \url{https://github.com/asbroad/model_based_shared_control}, and include a video depicting a user's control of the dynamic system and the impact of model-based shared control in the supplementary material.

\begin{acks}
This material is based upon work supported by the National Science Foundation under Grant CNS 1329891. Any opinions, findings and conclusions or recommendations expressed in this material are those of the authors and do not necessarily reflect the views of the aforementioned institutions.
\end{acks}

\bibliographystyle{SageH}
\bibliography{references.bib}

\end{document}